\documentclass[letterpaper]{article} 
\usepackage{aaai23}  
\usepackage{times}  
\usepackage{helvet}  
\usepackage{courier}  
\usepackage[hyphens]{url}  
\usepackage{graphicx} 
\usepackage{natbib}  
\usepackage{caption} 
\usepackage{algorithm}
\usepackage{algorithmic}

\usepackage{newfloat}
\usepackage{listings}
\DeclareCaptionStyle{ruled}{labelfont=normalfont,labelsep=colon,strut=off} 
\floatstyle{ruled}
\newfloat{listing}{tb}{lst}{}
\floatname{listing}{Listing}
\usepackage{tabularx}
\usepackage{booktabs}
\usepackage{multirow}
\usepackage{microtype}      
\usepackage{amsthm}
\usepackage{amssymb}
\usepackage{amsmath}

\usepackage{subcaption} 

\newtheorem{definition}{Definition}
\newtheorem{lemma}{Lemma}

\DeclareMathOperator*{\argmin}{arg\,min}

\newcommand{\subalign}[1]{
  \vcenter{
    \baselineskip 0pt
    \lineskiplimit 0pt \lineskip 0pt
    \ialign{&$\scriptstyle{}##$ & $\scriptstyle{}##$\cr
      #1\crcr
    }
  }
}

\newcommand{\ourmodel}{PHN-HVI }

\title{Improving Pareto Front Learning via Multi-Sample Hypernetworks}
\author {
    Long P. Hoang\textsuperscript{\rm 1,\rm 2\footnote{The author started this paper as part of the graduation thesis.}},
    Dung D. Le\textsuperscript{\rm 1},
    Tran Anh Tuan\textsuperscript{\rm 2},
    Tran Ngoc Thang\textsuperscript{\rm 2}
}
\affiliations {
    \textsuperscript{\rm 1} College of Engineering and Computer Science, VinUniversity  \\
    \textsuperscript{\rm 2} School of Applied Mathematics and Informatics, Hanoi University of Science and Technology \\
    long.hp@vinuni.edu.vn, dung.ld@vinuni.edu.vn, tuan.ta181295@sis.hust.edu.vn, thang.tranngoc@hust.edu.vn 
}

\begin{document}
\maketitle


\begin{abstract}
 Pareto Front Learning (PFL) was recently introduced as an effective approach to obtain a mapping function from a given trade-off vector to a solution on the Pareto front, which solves the multi-objective optimization (MOO) problem. Due to the inherent trade-off between conflicting objectives, PFL offers a flexible approach in many scenarios in which the decision makers can not specify the preference of one Pareto solution over another, and must switch between them depending on the situation. However, existing PFL methods ignore the relationship between the solutions during the optimization process, which hinders the quality of the obtained front. To overcome this issue, we propose a novel PFL framework namely \ourmodel, which employs a hypernetwork to generate multiple solutions from a set of diverse trade-off preferences and enhance the quality of the Pareto front by maximizing the Hypervolume indicator defined by these solutions. The experimental results on several MOO machine learning tasks show that the proposed framework significantly outperforms the baselines in producing the trade-off Pareto front.
\end{abstract}

\section{Introduction}
\begin{figure*}[!htb]
    \centering
    \includegraphics[width=.95\textwidth]{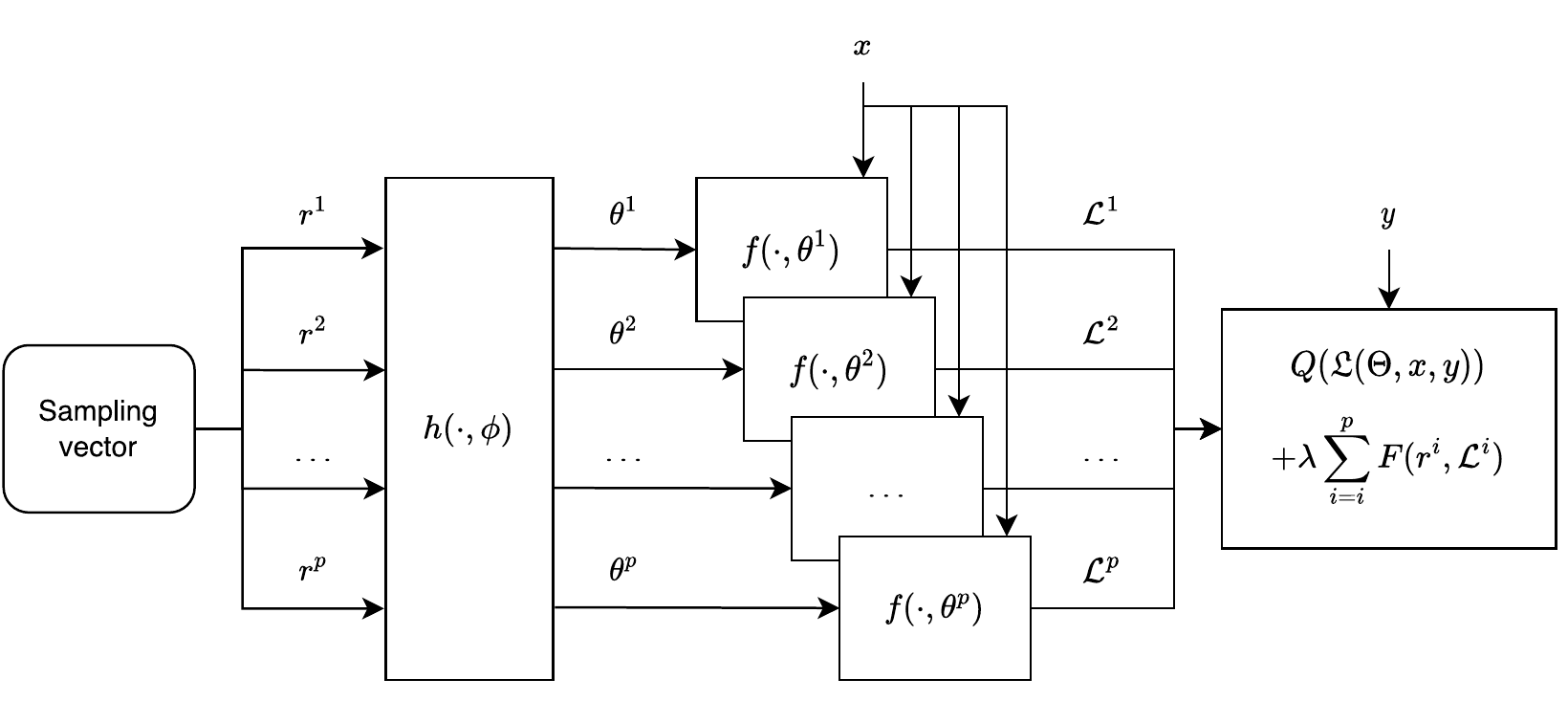}
    \caption{Multi-Sample Hypernetwork framework}
\end{figure*}
Multi-objective optimization has been shown prevalent in many machine learning applications with conflicting objectives such as in computer vision \citep{cv3, cv2, anh2022multi}, speech \& natural language processing \citep{sn1,sn3}, and recommender system \citep{rs2,rs3, le2020stochastically, le2021efficient}, etc. \cite{sener2019multitask} states that multi-task learning can be formulated as a multi-objective optimization (MOO) problem, in which the optimal decisions need to be taken in the presence of trade-offs between two or more conflicting objectives. Typically in MOO problems, the set of optimal solutions is called Pareto front, in which each solution on the front represents different trade-off between the objectives. However, most of the recent MOO algorithms must know the trade-off in advance or require a separate model to be trained for each point on the Pareto front \cite{navon2021learning}. This limits the flexibility as there are many situations in which the user can only make a trade-off decision on the fly, such as selecting between the shortest but congested route and the farther but more sparse one, or adapting the investment strategy when the marketplace changes. 

For that reason, \cite{controllable, navon2021learning, ruchte2021scalable} recently explore and develop a research direction called Pareto Front Learning (PFL), which attempts to approximate the entire Pareto front. However, these works only update the weights in the optimization algorithms by a random preference vector at each iteration, bringing with them restrictions in MOO that render them inappropriate for the PFL problem (see Section \ref{st:5} for more details), and ignoring the dynamic relationship between the Pareto optimal solutions. We argue that this information is crucial in producing a high-quality Pareto front because having a variety of solutions representing different trade-off preferences between the objective functions offers us a global view of the current Pareto front. Therefore, we propose to use hypervolume indicator to guide the optimization process. Specifically, we build upon the work of \cite{navon2021learning} to allow the derivation of multiple Pareto optimal solutions from multiple preference vectors, followed by the hypervolume (defined by the obtained solutions) maximization step to improve the learnt Pareto front.  

{\bf Contributions.} Our contributions can be summarized in the following:
\begin{itemize}
    \item \emph{Firstly,} we provide a mathematical formulation of PFL with Hypernetwork, which originally introduced in \cite{navon2021learning}.
    \item \emph{Secondly,} we propose a novel framework, namely \ourmodel that improves PFL problem using multi-sample hypernetworks and hypervolume indicator maximization.
    \item \emph{Thirdly,} we conduct comprehensive experiments on several multi-task learning datasets to validate the effectiveness of \ourmodel compared to baseline methods.   
\end{itemize}

{\bf Organization.} Section \ref{st:2} provides the background knowledge for the PFL problem. In Section \ref{st:3}, we describe our framework \ourmodel and its efficient variant with the partitioning algorithm. In Section \ref{st:4}, we review related work in the literature and briefly discuss both of their accomplishments and shortcomings. In Section \ref{st:5}, we conduct experiments from comparison and provide a model behavior analysis. We conclude the paper in Section \ref{st:6} and discuss possible research directions as future work.

\section{Preliminary}
\label{st:2}
Multi-task learning seeks to find $\theta^* \in \Theta$ to optimize $J$ loss functions as described in the following: 
\begin{align}
    \theta^* &= \argmin_{\theta} \mathbb{E}_{(x, y) \sim p_D} \mathcal{L}\left(y, f\left(x; \theta \right)\right)\\
    \mathcal{L}\left(y, f\left(x; \theta \right)\right) &=\left\{ \mathcal{L}_1\left(y, f\left(x; \theta \right)\right),  
    \nonumber
    \dots,  
     \mathcal{L}_J\left(y, f\left(x; \theta \right)\right) \right\}
\end{align}
in which, $p_D$ denotes the data distribution,  $\mathcal{L}_j\left(\cdot, \cdot \right):
\mathcal{Y} \times \mathcal{Y} \rightarrow \mathbb{R}_{>0}, \forall j \in 
\left\{1, \dots, J\right\}$ denotes the $j-th$ loss function, and $f\left(x; \theta \right): \mathcal{X} \times \vartheta \rightarrow \mathcal{Y}$ denotes the neural network with parameter $\theta$.

\begin{definition}[Dominance] A solution $\theta^a$ is said to dominate another solution $\theta^b$ if $$\mathcal{L}^j(y, f(x, \theta^a)) \leq \mathcal{L}^j(y, f(x, \theta^b)), \forall j \in \{1,...,J\}$$ and $\mathcal{L}(y, f(x, \theta^a)) \neq \mathcal{L}(y, f(x, \theta^b))$. We denote this relationship as $\theta^a \prec \theta^b$.
\label{def:paretodominance}
\end{definition}

\begin{definition}[Pareto optimal solution] A solution $\theta^a$ is called Pareto optimal solution if there exists no $\theta^b$ such as $\theta^b \prec \theta^a$.
\label{def:paretooptimality}
\end{definition}

\begin{definition}[Pareto front] The set of Pareto optimal is Pareto set, denoted by $\mathcal{P}$, and the corresponding images in objectives space are Pareto front $\mathcal{P}_f=\mathcal{L}(y, f(x, \mathcal{P}))$.
\end{definition}

In \cite{navon2021learning}, the authors introduced the term Pareto Front Learning. Here we provide a mathematical formulation of Pareto Front Learning with Hypernetwork, which sets the basis for our proposed framework in Section \ref{st:3}:
\begin{definition}[PFL with Hypernetwork]
Pareto Front Learning is a one-shot optimization approach to approximate the entire Pareto front by solving problem:
\begin{equation}
\begin{aligned}
    & \phi^* = \argmin_{\phi \in \mathbb{R}^n} \mathbb{E}_{r \sim p_{\mathcal{S}^J}, (x, y) \sim p_D} \;F(\mathcal{L}(y,f(x, \theta^r)), r) \\
    & \text{ s.t. } \theta^r = h(r, \phi) \in \mathcal{P}, h(\Omega, \phi^*) = \mathcal{P}
    \label{eq:plf}
\end{aligned}
\end{equation}
where $h: \mathcal{S}^J \times \Phi \rightarrow \Theta$, random variable $r$ is the preference vector that formulates a trade-off between loss functions, $\mathcal{S}^J = \{\lambda \in \mathbb{R}^J_{>0}: \sum_j \lambda_j = 1\}$ is the set feasible values of random variable $r$ and $p_{\mathcal{S}^J}$ is a random distribution on $\mathcal{S}^J$ and $F(\cdot, \cdot): \mathcal{S}^J \times \mathbb{R}^J \rightarrow \mathbb{R}$ is an extra criterion function, which helps us map a given preference vector with a Pareto optimal solution.

\end{definition}

\section{Multi-Sample Hypernetwork}
\label{st:3}
Multi-Sample Hypernetwork samples $p$ preference vectors $r^i$ and use $h(\cdot, \phi)$ to generate $p$ target networks $f(\cdot, \theta^{i}), i\in \{1,\dots, p\}$ where $\theta^{i} = h(r^i, \phi)$. Denote $\mathcal{L}^i =(\mathcal{L}_1(y, f\left(x; \theta^{i} \right)), \dots, \mathcal{L}_J(y, f\left(x; \theta^{i} \right)))$ is the vector of loss values, and $\mathfrak{L}(\Theta,x, y)=\left[\mathcal{L}^1, \dots, \mathcal{L}^p \right]$. Choose $p_{\mathcal{S}^J}$ as $\text{Dir}(\alpha)$. Different from \cite{navon2021learning}, Multi-Sample Hypernetwork is designed to solve: 
\begin{align}
    \label{eq:hpn-hvi}
    \min_{\phi} \;\nonumber \mathbb{E}_{\subalign{ r^i\sim \text{Dir}(\alpha) \\(x,y) \sim p_D}} \;\;\;\;\;\; Q(\mathfrak{L}(\Theta,x, y)) + \sum_{i=1}^p F(r^i, \mathcal{L}^i)
\end{align}
where $Q: \mathbb{R}^J \rightarrow \mathbb{R}$ is a monotonically decreasing function, meaning $\forall \theta^b \in \Theta^B \exists \theta^a \in \Theta^A$ : $\theta^a \prec \theta^b  \Rightarrow Q(\mathfrak{L}(\Theta^A,x, y)) < Q(\mathfrak{L}(\Theta^B,x, y))$ or $\nabla Q(\mathfrak{L}(\Theta,x, y))$ create a Pareto improvement, and $F: \mathcal{S}^J \times \mathbb{R}^J \rightarrow \mathbb{R}$ is a complete function, for  example, $\{\argmin_{\mathcal{L} \in \mathcal{P}_f} F(r, \mathcal{L}): \forall r \in \mathcal{S}^J\} = \mathcal{P}_f$.

\subsection{HV Indicator Hypernetwork}
HV Indicator Hypernetwork (\ourmodel) solves the problem:
\begin{flalign}
    \min_\phi - \mathbb{E}_{\subalign{ &r^i\sim \text{Dir}\left(\alpha \right)\nonumber\\&(x,y) \sim p_D}} \mathrm{HV}\left(\mathfrak{L}(\Theta; x, y)\right) + \lambda \sum_{i=i}^p \cos(\vec{r^i}, \vec{\mathcal{L}^i}) \text{ } \textit{ } \textbf{ }
\end{flalign}

\begin{lemma}[\cite{hv2}]
If $\mathrm{HV}\left(\mathfrak{L}(\Theta; x, y)\right)$ is maximal, then $\theta^{i} \in \mathcal{P}, i \in \{1,\dots,p\}$. 
\end{lemma}
The Hypervolume (HV) function will maximize the hypervolume of the Pareto front, pushing the Pareto front approximated by hypernetwork to the truth Pareto front. Meanwhile, the cosine similarity penalty function will spread the Pareto front and get the Pareto optimal solution close to the given preference vector. It is clear that cosine similarity is a function with a complete property, HV is a monotonic function.

Using hypervolume gradient algorithm of the independent $p$ neural networks described in \cite{hypervolumegrad}, we obtain the following, which is scalable for Domination-Ranked Fronts \citep{Dominated-front1, Deist_2020}:
\begin{equation}
\begin{aligned}
& \nabla_{\mathcal{L}^i} \left[-\mathrm{HV}\left(\mathfrak{L}(\Theta; x, y)\right) \right ] \\
&= \left( -\frac{\partial \mathrm{HV}\left(\mathfrak{L}(\Theta; x, y)\right)}{\partial \mathcal{L}_1(y, f(x, \theta^{i}))},\dots, -\frac{\partial \mathrm{HV}\left(\mathfrak{L}(\Theta; x, y)\right)}{\partial \mathcal{L}_p(y, f(x, \theta^{i}))} \right)
\label{eq:1}
\end{aligned}
\end{equation}
We can approximate the descent direction of HV function for parameter $\phi$ based on \eqref{eq:1} by: 
\begin{equation}
\begin{aligned}
d = -\sum_{i=1}^p \sum_{j=1}^J \frac{\partial \mathrm{HV}\left(\mathfrak{L}(\Theta; x, y)\right)}{\partial \mathcal{L}_j(\theta^{i}, f(x, \theta^{i}))}\frac{\partial \mathcal{L}(\theta^{i}, f(x, \theta^{i}))}{\partial \phi}
\label{eq:6}
\end{aligned}
\end{equation}
The final update direction of HV Indicator Hypernetwork is:
        \begin{align}
              d_{update}
              = d - \lambda\sum_{i=1}^p \frac{\partial}{\partial \phi}\left( \frac{\vec{r^i}\vec{\mathcal{L}}(y, f\left(x; \theta^{i} \right))}{\|\vec{r^i}\|\|\vec{\mathcal{L}}(y, f\left(x; \theta^{i} \right))\|}\right) 
        \end{align}
It should be noted that $\phi$ is the only parameter for \ourmodel that has to be optimized. On the inference phase, \ourmodel just needs one vector $r$  to provide a corresponding Pareto local optimum $\theta^r$.

\subsection{Partitioning Algorithm}
\label{sst:partition}
It is necessary to sample the preference vectors $r^1,\dots, r^p$ to cover the objective space evenly to make HV and cosine similarity interact well. This is quite simple in 2D space using $p+1$ vectors $\{\cos(\frac{i\pi}{2p}, \sin(\frac{i\pi}{2p}) \}$ to partition the object space into $p$ subregions $\Omega^i, i=1,...p$ and randomly sample the vector $r^i$ of subregion $\Omega^i$. However, how to take a partitioned random sample in any dimensional $J$ space? A good partition with $N$ subregions must satisfy:
\begin{align}
    \cup_{i=1}^N (\Omega^i)= S^J \text{ and }  \Omega^{i'} \cap_{i' \neq i} \Omega^i = \emptyset
\end{align}
Base on \cite{das}, we define subregions $\Omega^i$ by points $u=(u_1, \dots, u_J) \in U \subset \mathcal{S}^J$ such that:
\begin{align}
    u_1 \in \{0, \delta, 2\delta, \dots, ..., 1\} \text{ s.t }  \frac{1}{\delta}=k \in \mathbb{N}^* 
\end{align}
If $1<j<J-1$, $m_i = \frac{\delta}{u_i}, 1 \leq i<j-1$, we have:
\begin{align}
    u_j \in \{0, \delta, \dots, (p-\sum_{i=1}^{j-1}m_i)\delta\} \text{, }  u_J = 1 - \sum_{j=1}^{J-1}u_j
\end{align}
\begin{figure}[!htb]
    \centering
    \includegraphics[width=0.8\columnwidth]{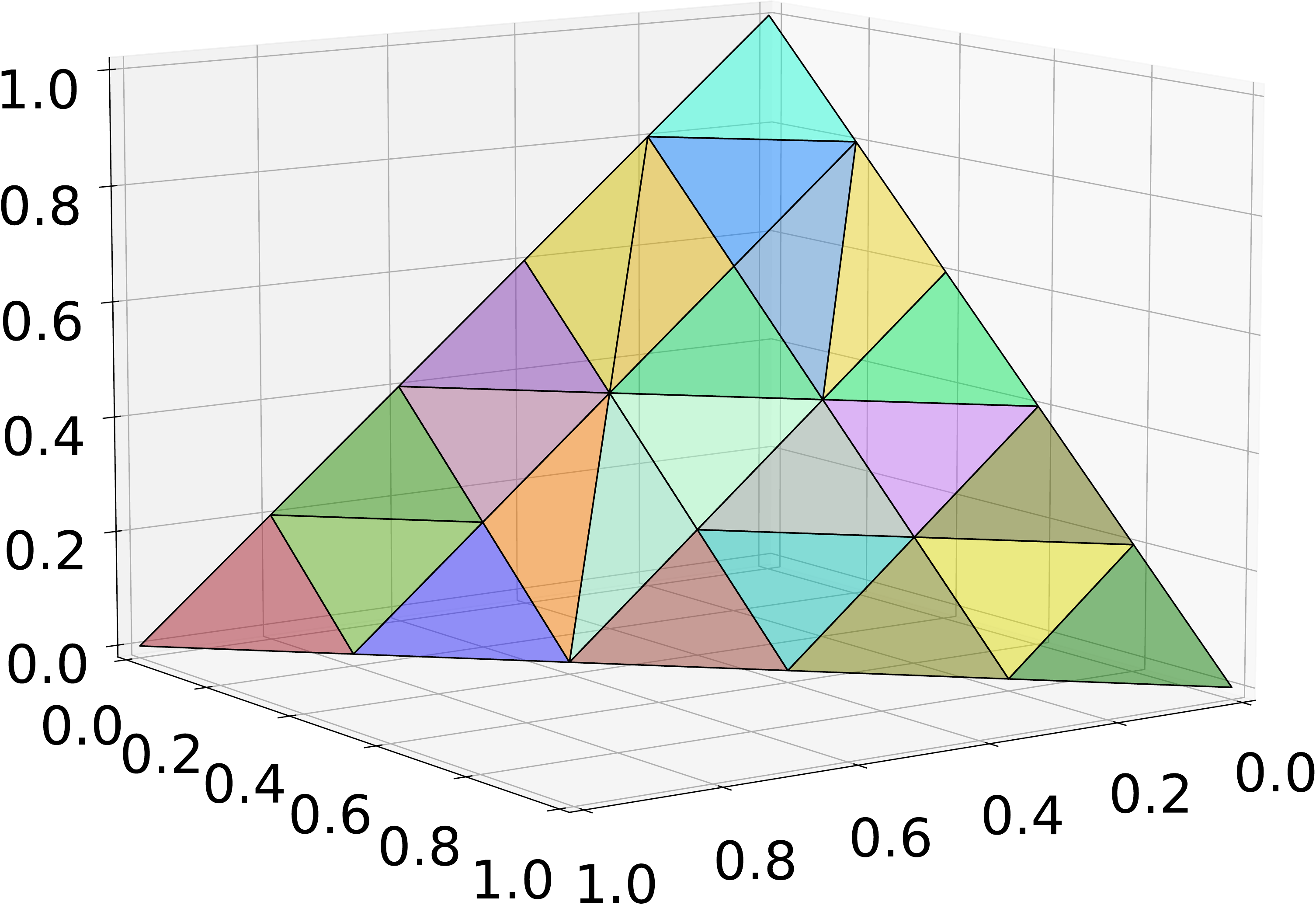}
    \caption{Partitioning algorithm with $J=3, \delta=0.2$}
    \label{fig:partition_result}
\end{figure}
Using the Delaunay triangulation algorithm \cite{Edelsbrunner1985VoronoiDA} for these points, we obtain the required partition. Figure \ref{fig:partition_result} is the illustrative result of the algorithm in 3D space. However, the number of points $u \in U$ in the space $\mathbb{R}^J$ and $k = \frac{1}{\delta}$ is $\Big (^{J+k-1}_k \Big)$. Therefore, we will not be able to freely set the number of rays $p$ for the HV Indicator Hypernetwork, and the number of partitions also increases exponentially as $J$ and $k$ increase, resulting in a huge amount of computation. Therefore, we will only use the partition sampling algorithm for the 2-objective optimization problem and still randomly sample the preference vectors $r^1, \dots, r^p \sim \text{Dir}(\alpha)$ with optimization problems of 3 or more objective functions.

\begin{algorithm}
\caption{\ourmodel optimization algorithm}
\begin{algorithmic}[1]
\WHILE{not converged}
\IF{$J=2$}
\STATE $r^1, \dots, r^p \sim \text{partition\_sampling}$
\ELSE
\STATE $r^1, \dots, r^p \sim \text{Dir}(\alpha)$
\ENDIF
\STATE Compute $\left[\theta^{i} := h(r^i, \phi)\right]_{i=1}^p$ \vspace{5pt}
\STATE Compute $d$ by Equation (\ref{eq:6})

\STATE $d_{update} := d - \lambda\sum_{i=1}^p \frac{\partial}{\partial \phi}\left( \frac{\vec{r^i}\vec{\mathcal{L}}(y, f\left(x; \theta^{i} \right))}{\|\vec{r^i}\|\|\vec{\mathcal{L}}(y, f\left(x; \theta^{i} \right))\|}\right) $ 
\STATE $\phi \xleftarrow{} \phi - \eta \times d_{update} $
\ENDWHILE
\RETURN $\phi$
\end{algorithmic}
\label{alg:1}
\end{algorithm}

\subsection{Stein Variational Hypernetwork}
Beside Hypervolume maximization, it is also possible to integrate the multi-sample hypernetwork with Stein variational gradient descent \cite{moosvgd} for profiling the Pareto front. Similar to \ourmodel, Stein variational hypernetwork generates $p$ target networks. 

Denote $g(\phi, r^i)$ is min norm vector in the convex hull $\mathcal{CH}_{\mathcal{L}^i}$ of the gradients $\nabla_{\phi}\mathcal{L}^i$:
\begin{align}
    g(\phi, r^i) = \argmin_{g \in \mathcal{CH}_{\mathcal{L}^i}} \|g\|^2
\end{align}
We can approximate the update direction for an input trade-off vector $r^i$ as follows:
\begin{align}
    d^i = \sum_{i'=1}^p \left[ g(\phi, r^{i'}) k(\mathcal{L}^i, \mathcal{L}^{i'}) - \nabla_{\phi}k(\mathcal{L}^i, \mathcal{L}^{i'}) \right]
\end{align}
where $k(\mathcal{L}^i, \mathcal{L}^{i'})=\text{det}(2\pi \sigma^2I)^{-\frac{1}{2}}\exp(-\frac{1}{2\sigma^2}\|\mathcal{L}^i-\mathcal{L}^{i'}\|^2)$.
The update direction of the hypernetwork's parameter is:
\begin{align}
    d_{update} = \sum_i^p d^i - \lambda\sum_{i=1}^p \frac{\partial}{\partial \phi}\left( \frac{\vec{r^i}\vec{\mathcal{L}}(y, f\left(x; \theta^{i} \right))}{\|\vec{r^i}\|\|\vec{\mathcal{L}}(y, f\left(x; \theta^{i} \right))\|}\right) 
\end{align}

This demonstrates the flexibility of using Multi-Sample Hypernetworks to enable the iterative update of a set of solution points simultaneously to push them toward the Pareto front. However, we reserve the investigation of this aspect for future work. 

\section{Related Work}
\label{st:4}
\indent \textbf{Multi-Objective Optimization.} Rather than a single solution, MOO aims to find a set of Pareto optimal solutions with different trade-offs. For small-scale multi-objective optimization problems, genetic algorithms such as VEGA \cite{inproceedings}, NSGA-III \cite{6600851}, etc. are popular methods for finding a set of well-distributed Pareto optimal solutions in a single run. Scalarization algorithms (the Tchebysheff method \cite{article} and its variants) use weighted functions to transform several objectives into a single objective, but, these methods need convex function conditions to approximate the entire Pareto front. Therefore, \cite{thang2020monotonic} used a monotonic optimization approach to obtain an approximation of the weakly Pareto optimal set for solving strictly quasiconvex multi-objective programming, a general case of convex multi-objective programming. Another approach, as developed by \cite{gradientbased1}, \cite{gradientbased3} find a common descent direction of all objectives at each iteration, as a result, they cannot take in decision-makers preferences on the problem.\\
\indent \textbf{Multi-Task Learning.} The use of MOO in machine learning, particularly for Multi-Task Learning, is not new.  Some algorithms approach task balancing, such as \cite{kendall2018multitask} (homoscedastic uncertainty), \cite{chen2018gradnorm} (balance learning), etc. \cite{sener2019multitask} used Frank-Wofle algorithm to solve the constrained optimization problem of MGDA \cite{gradientbased3}, while \cite{yu2020gradient} corrects the direction of conflict gradients. Due to their approach to balanced solutions, these methods are not suitable for simulating trade-offs between objectives. \\
Using preference vectors, \cite{lin2019pareto} may identify a variety of solutions by breaking down a multitask learning problem into several subproblems of many subspaces, whereas \cite{pmlr-v119-mahapatra20a} can precisely identify the Pareto optimal solution belonging to inverse ray.  \cite{ma2020efficient} has the ability to locate Pareto optimal solutions close to a given Pareto optimal solution.
The Pareto front with finite points approximated by several neural networks is another research direction. \cite{deist2021multiobjective} has done this using Hypervolume Maximization. The techniques for calculating the gradient of the hypervolume were created by \cite{hypervolumegrad}, and \citep{Dominated-front1, Deist_2020} further improved them. To increase the entropy of these points, \cite{moosvgd} combined MGDA \cite{gradientbased3} with Stein Variational Gradient Descent \cite{svgd}. Unfortunately, as the number of preference vectors and objective functions rises, the computational cost of these algorithms becomes impractical. \\
\indent \textbf{Hypernetwork and Pareto Front Learning.} Hypernetwork, which generates weights for other networks (target networks), can be applied to large-scale problems by using chunking \cite{hyper_origin}. The result in \cite{controllable} is poor since they constructed a hypernetwork but did not explicitly map a preference vector with a Pareto optimal solution. \cite{navon2021learning} presented PHN-LS and PHN-EPO, which combine optimization techniques like Linear Scalarization (LS) and EPO \cite{pmlr-v119-mahapatra20a} with hyperwork, to approximate the complete Pareto front in a single training. As a result, the two models still contain the features and drawbacks of these techniques.  In order to tackle PFL, \cite{ruchte2021scalable} combined preference vectors with data using a single network and used the LS loss function together with cosine similarity. However, it is still unclear how to concatenate preference vectors with arbitrary data, and not all situations allow for efficient interaction between LS and cosine similarity.

\section{Experiments}
\label{st:5}
\begin{figure*}
    \centering
    \includegraphics[width=\textwidth]{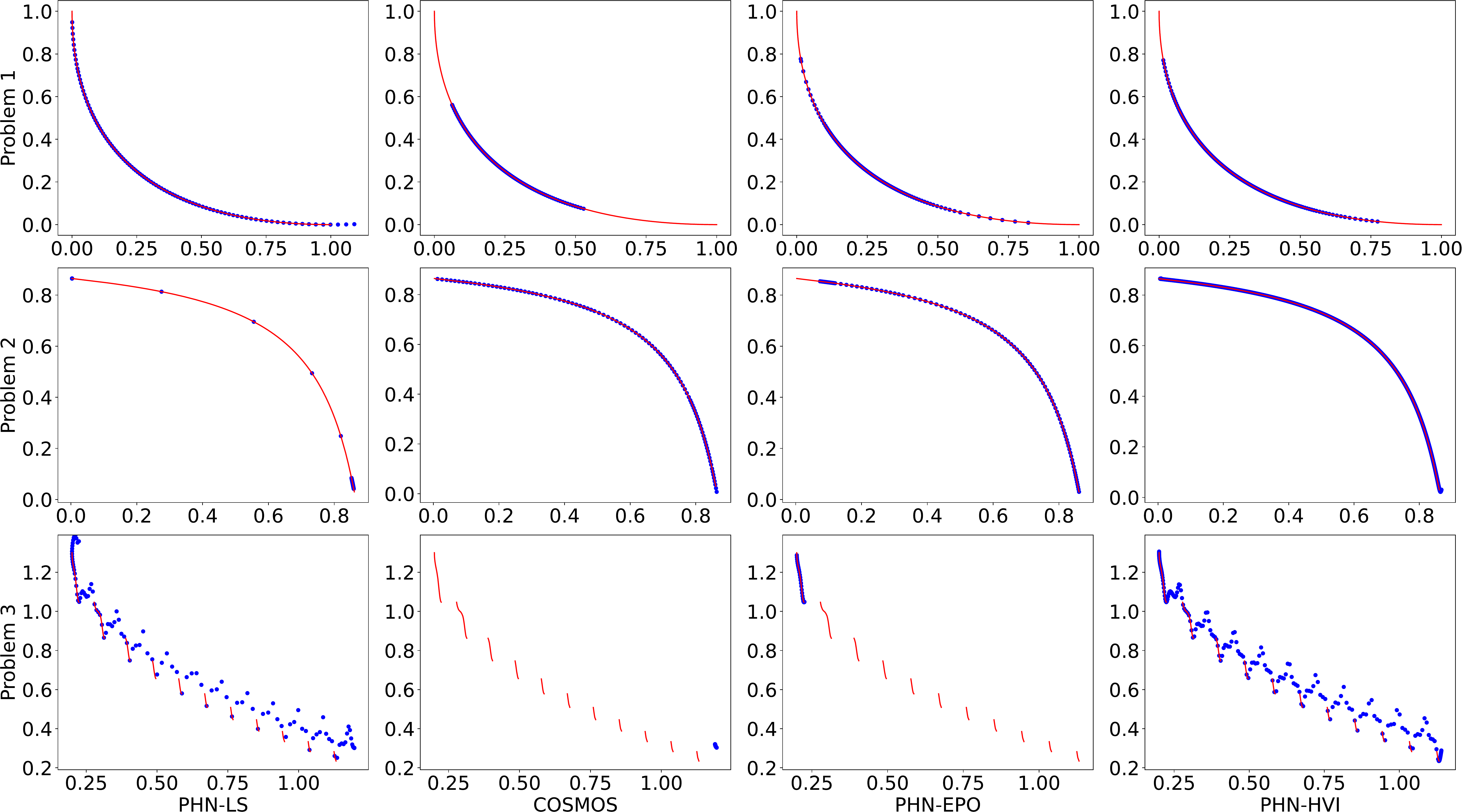}
    \caption{Toy 2D plots. The red curves are truth Pareto front. The blue points are approximate Pareto fronts}
\end{figure*}

\begin{figure*}[!htb]
     \centering
     \begin{subfigure}[b]{0.24\textwidth}
         \centering
         \includegraphics[width=\textwidth]{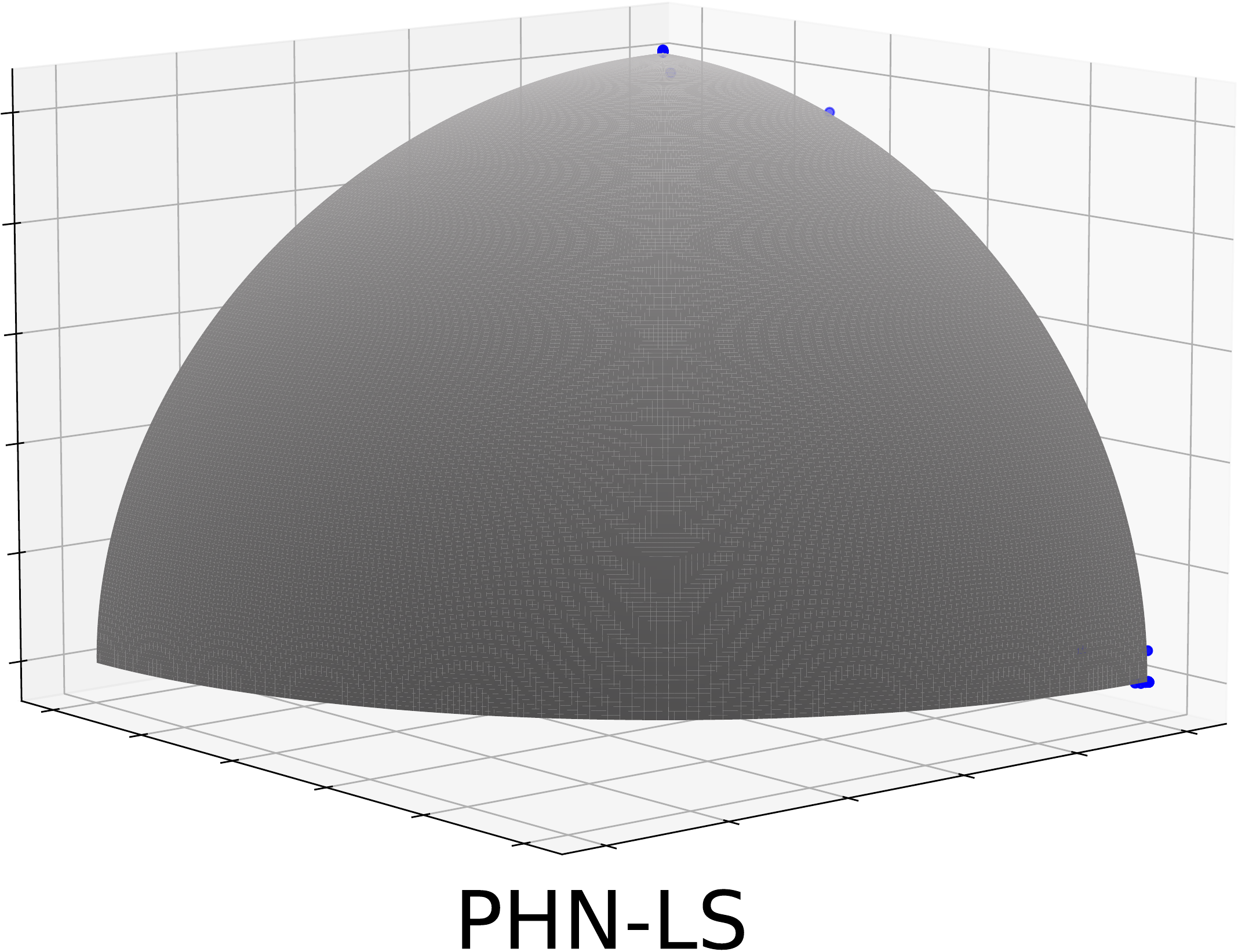}
     \end{subfigure}
     \hfill
     \begin{subfigure}[b]{0.24\textwidth}
         \centering
         \includegraphics[width=\textwidth]{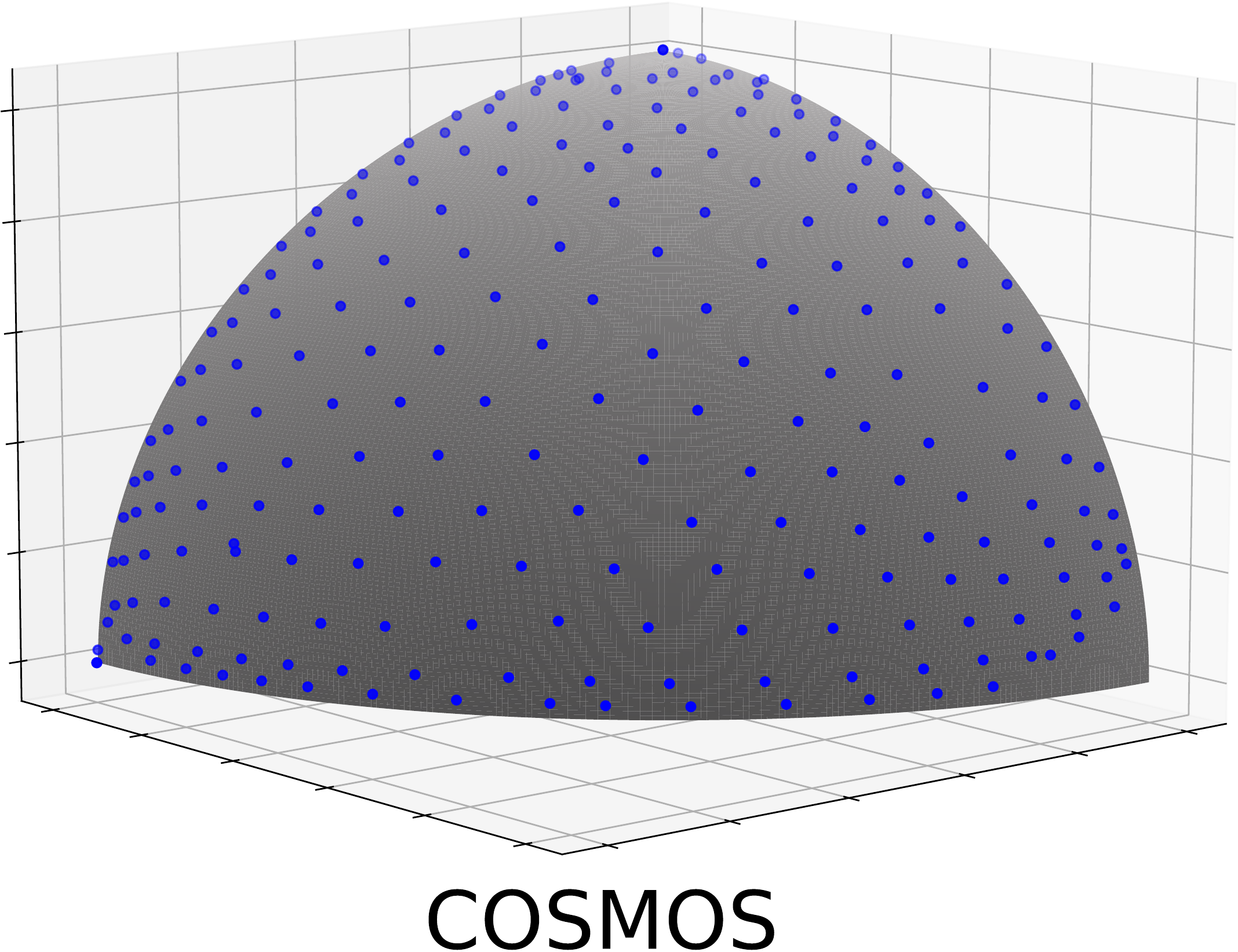}
     \end{subfigure}
     \hfill
     \begin{subfigure}[b]{0.24\textwidth}
         \centering
         \includegraphics[width=\textwidth]{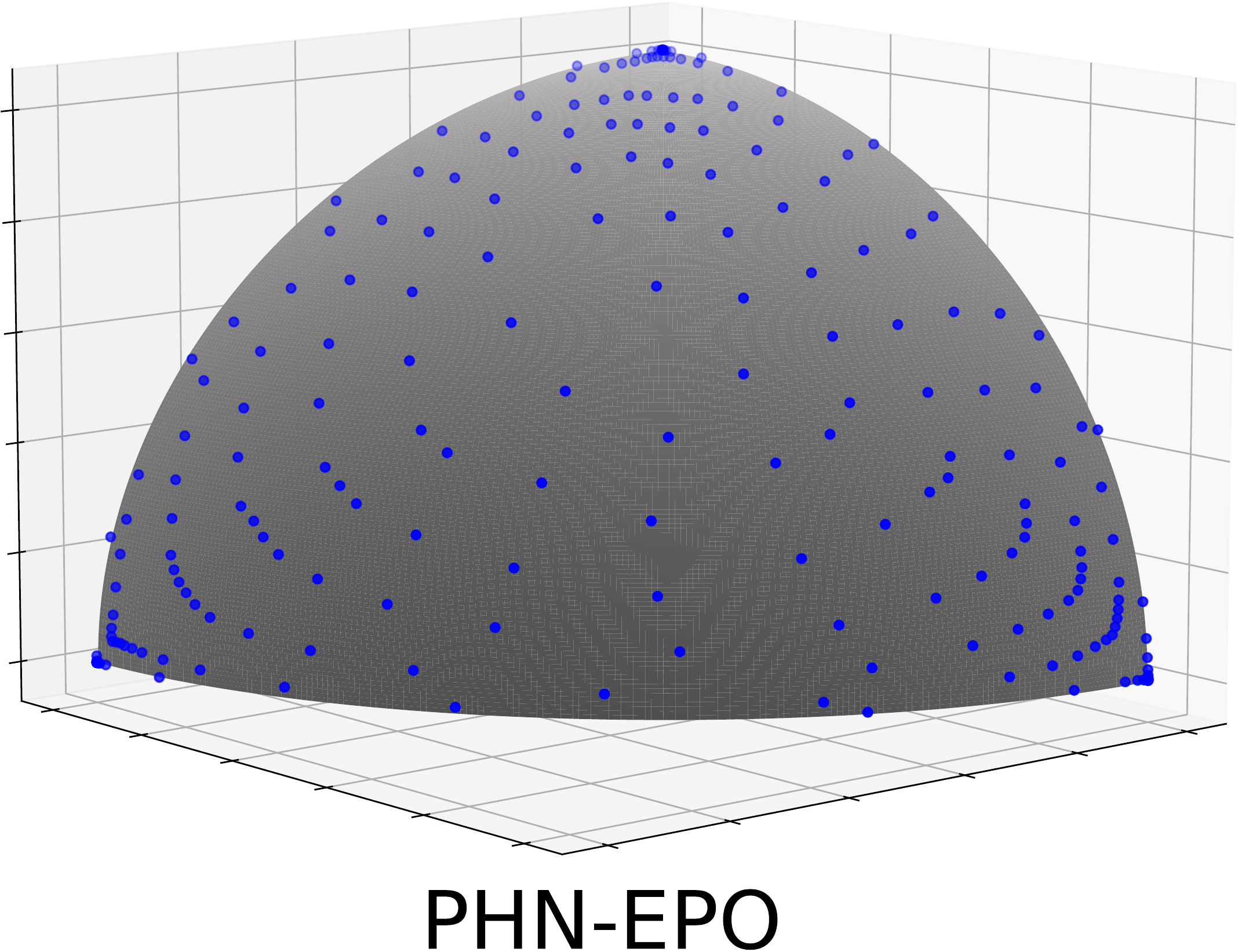}
     \end{subfigure}
     \hfill
     \begin{subfigure}[b]{0.24\textwidth}
         \centering
         \includegraphics[width=\textwidth]{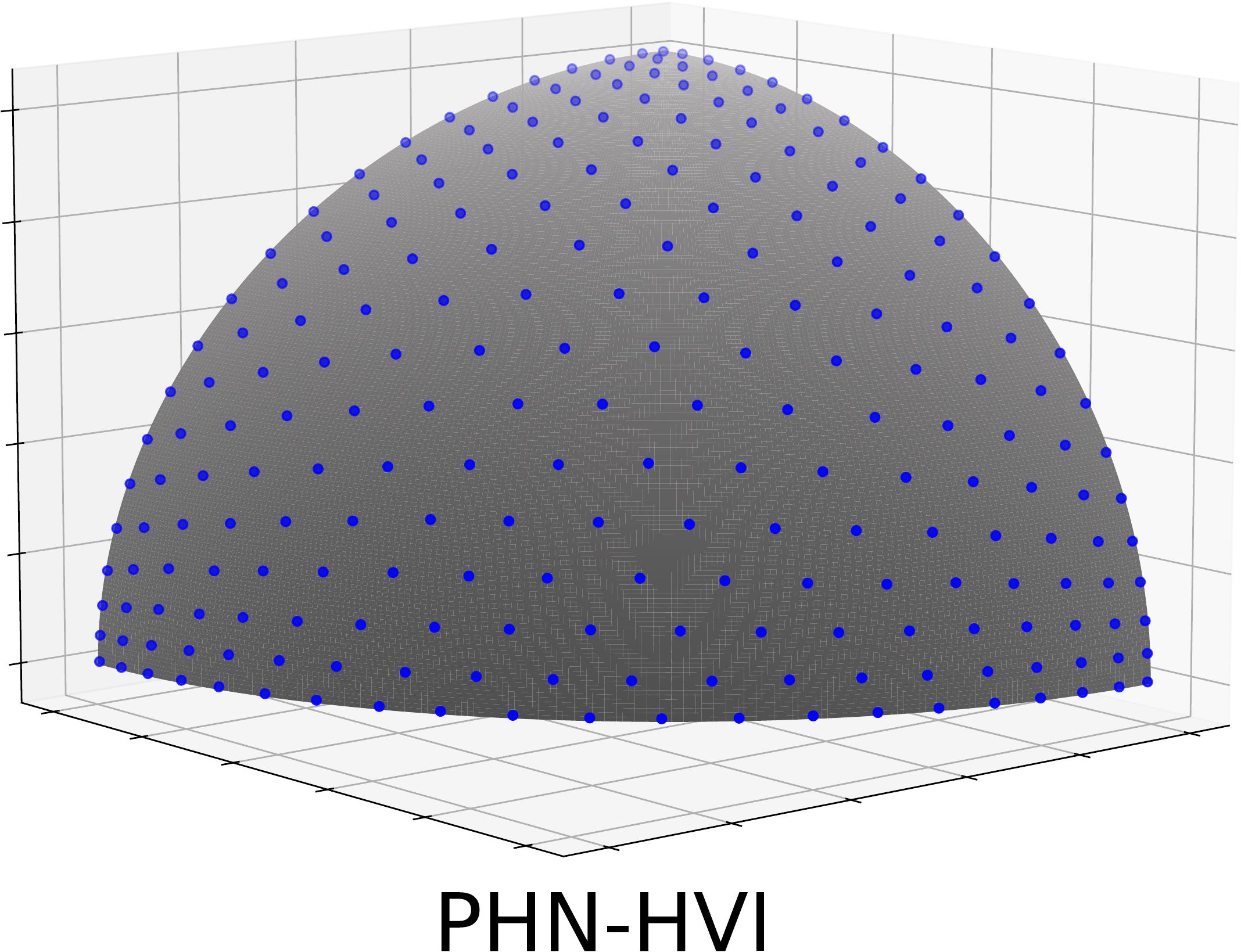}
     \end{subfigure}
      \caption{Toy 3D plots}
      \label{fig:3d}
\end{figure*}

For all methods base on hypernetwork, we use a feed-forward network with various outputs to parameterize $h(\phi, r)$. The target network's weight tensor is produced by each output in a distinct way. Specifically, the input $r$ is first mapped using a multi-layer perceptron network to a higher dimensional space in order to create shared features. A weight matrix for each layer in the target network is created by passing these features across fully connected layers. Experiments demonstrate that \ourmodel outperforms other methods.

\textbf{Baselines:} We compare \ourmodel\footnote{We publish the code at \url{https://github.com/longhoangphi225/MultiSample-Hypernetworks}} with the current state-of-the-art of PFL methods: \textbf{PHN-LS}, \textbf{PHN-EPO} \cite{navon2021learning}, and \textbf{COSMOS} \cite{ruchte2021scalable}.

\textbf{Evaluation Metric:} The area dominated by Pareto front is known as Hypervolume \cite{zitzler1999multiobjective}. The higher Hypervolume, the better Pareto front.  

\textbf{Training Settings:} Our target network has the same architecture as the baselines in all experiments. For toy examples, the optimization processes are run for 10,000 iterations, and 200 evenly distributed preference vectors are used for testing. On multi-task problems, the dataset is split into three subsets: training, validation, and testing. The model with the highest HV in the validation set will be evaluated. All methods are evaluated by the same well-spread preference vectors.

All experiments and methods in this paper are implemented with Pytorch \cite{pytorch} and trained on a single NVIDIA GeForce RTX3090.
\subsection{Toy Examples}
\label{sst:toy}
In this section, we will investigate the quality of the Pareto front generated by \ourmodel and the PFL baselines using the following toy examples:

\noindent {\bf Problem 1 \cite{moosvgd}}
\begin{equation}
  \begin{aligned}
    \mathcal{L}_1(\theta) = (\theta)^2 \text{ , } \mathcal{L}_2(\theta) = (\theta-1)^2, \text{ s.t. }\theta \in \mathbb{R}
\end{aligned}  
\end{equation}

\noindent {\bf Problem 2 \cite{lin2019pareto}}
\begin{equation}
\begin{aligned}
    & \mathcal{L}_1(\theta) = 1 - \exp\{ -\| \theta - 1/\sqrt{d} \|^2_2 \}, \\
    & \mathcal{L}_2(\theta) = 1 - \exp\{ -\| \theta + 1/\sqrt{d} \|^2_2 \} \\
    & \text{s.t }\theta \in \mathbb{R}^d, d=100
\end{aligned}
\label{pb1}
\end{equation}
\noindent {\bf Problem 3}
\begin{align}
    & \mathcal{L}_1(\theta) = \cos^2(\theta_1) + 0.2, \\
    & \mathcal{L}_2(\theta) = 1.3 + \sin^2(\theta_2) - \cos(\theta_1) - 0.1\sin^5(22 \pi \cos^2\theta_1) \nonumber \\
    & \text{s.t }\theta \in \mathbb{R}^2 \nonumber
\end{align}
{\bf Problem 4 (DTLZ4 with 3 Objective Functions)}
\begin{align}
    & \mathcal{L}_1(\theta) = \cos(\theta_1 \frac{\pi}{2})\cos(\theta_2 \frac{\pi}{2})(\sum_{i=3}^{10} (\theta_i -0.5)^2 +1), \\
    & \mathcal{L}_2(\theta) = \cos(\theta_1 \frac{\pi}{2})\sin(\theta_2 \frac{\pi}{2})(\sum_{i=3}^{10} (\theta_i -0.5)^2 +1), \nonumber \\
    & \mathcal{L}_3(\theta) = \sin(\theta_1 \frac{\pi}{2})(\sum_{i=3}^{10} (\theta_i -0.5)^2 +1), \nonumber \\
    & \text{s.t }\theta \in \mathbb{R}^{10}, 0 \leq \theta_i \leq 1 \nonumber
    \label{pb4}
\end{align}

\begin{table*}[!htb]
\centering
\begin{tabular}{| l | >{\centering\arraybackslash}m{2.2cm} | >{\centering\arraybackslash}m{2.25cm} | >{\centering\arraybackslash}m{2.25cm} | >{\centering\arraybackslash}m{1.cm} | >{\centering\arraybackslash}m{1.cm} | >{\centering\arraybackslash}m{1.5cm} | }
\toprule
             {\bf Method}    &  {\bf Multi-MNIST} & {\bf Multi-Fashion} & {\bf Fash.+MNIST} & {\bf Drug} & {\bf Jura} & {\bf SARCOS}  \\ \midrule  
PHN-EPO   & 2.868  & 2.238  & 2.815  & 1.226 & 0.933 & 0.932 \\ 
PHN-LS    & 2.859   & 2.219  & 2.764 & 1.208 & 0.932 & 0.934  \\ 
COSMOS    & 2.959  & 2.324  & 2.838 & NA & 0.933 & 0.830  \\ \midrule
\textbf{\ourmodel}    & \textbf{3.012}   & \textbf{2.408} & \textbf{2.967}  & \bf 1.294 & \bf 0.946 & \bf 0.949 \\
\bottomrule
\end{tabular}
\caption{Results compared to the state-of-the-art methods on Hypervolume}
\label{tab:results}
\end{table*}

\begin{figure*}[!htb]
    \centering
    \includegraphics[width=\textwidth]{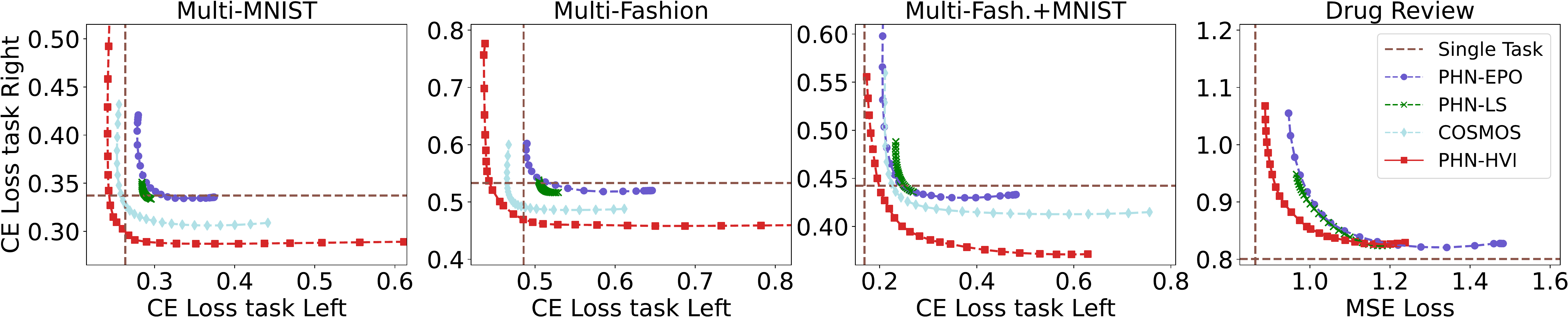}
    \caption{Pareto fronts are generated by methods}
    \label{fig:alltasks}
\end{figure*}

Since the objective function in Problem 1 is convex, practically all methods—in particular PHN-LS—perform well. Due to the concave form of the Pareto front in Problem 2, all of the solutions produced by PHN-LS gravitate to two ends, but COSMOS, a method that combines LS with consine similarity, can be effective in this situation. However, it is highly problematic in Problem 3 when LS and cosine similarity function are in direct conflict. \\
As a result of its disconnected Pareto front, Problem 3 presents a difficult challenge. In view of hypernetwork-based method, we have $\mathcal{L}_j(h(\phi, r))$ is a smooth function, because $\mathcal{L}_j(\theta)$ is smooth, and $h(\phi, r)$ is smooth (because $h$ is represented by a neural network). So on, if $\theta = h(\phi, r), \exists d_0, d_j, \epsilon_j \in \mathbb{R}_+, \forall r' \in \Omega: \|r' -r\| < d_0 \Rightarrow \theta' \in \mathcal{B}(\theta, d_i): \|\mathcal{L}_j(\theta') -\mathcal{L}_j(\theta)\| < \epsilon_i, j\in\{1,\dots, J\}$. Therefore as a consequence, in order to approximate the entire Pareto front, PHN-LS and \ourmodel must give some point that is not a Pareto optimal solution. And EPO Search, which combines uniformity function, common direction gradient descent, and a controlled increase or decrease in the objective function, prevents PHN-EPO from generating a non-Pareto Optimal solution, resulting in the generation of just a portion of the Pareto front. In this case, PHN-LS and \ourmodel are better than PHN-EPO, because they still can profile the Pareto front by removing the dominated solutions.\\
\indent We use the sigmoid function to address constraints in Problem 4. Using the sampling technique which is described in Section \ref{st:3}, we use 231 uniform 3D rays. In Figure \ref{fig:3d}, PHN-LS entirely fails, COSMOS and PHN-EPO offer solutions that are widespread but a little chaotic, while \ourmodel produces solutions that result in a pretty uniform distribution.

\subsection{Image Classification}
Three benchmark datasets—Multi-MNIST, Multi-Fashion, and Multi-Fashion+MNIST \cite{lin2019pareto} are used in our evaluation. For each dataset, we have a two-objective multitask learning problem that asks us to categorize the top-left (task left) and bottom-right (task right) items (task right). Each dataset has 20,000 test set instances and 120,000 training examples. 10\% of the training data are used for the validation split. Multi-LeNet \cite{sener2019multitask} is the network that all approaches aim to reach. We set $p=16, \lambda=5$ on the Multi-MNIST dataset, $p=16, \lambda=4$ on the Multi-Fashion dataset, and Multi-Fashion+MNIST dataset for \ourmodel.

In Quadrant I, we evaluate every methods using 25 uniformly distributed preference vectors. The results are shown in Table \ref{tab:results} and Figure \ref{fig:alltasks}. The HV was calculated by reference point (2, 2). The Pareto front of \ourmodel outperforms and covers the baselines entirely. The Pareto Front is generated by other techniques, especially PHN-LS, that are not widely dispersed. There are numerous solutions on the Pareto front of \ourmodel that may be achieved that are superior to a single task. 

\subsection{Text Classification and Regression}


In this investigation, we concentrated on the Drug Review dataset \cite{Felix2018}. This dataset comprises user evaluations of particular medications, details on pertinent ailments, and a user score that shows general satisfaction. We examine two tasks: (1) regression-based prediction of the drug's rating and (2) classify the patient's condition. 

This dataset consists of 215063 samples. 10\% of the data which has conditions with insufficient user feedback are removed.
Following that are 100 condition labels and 188155 samples. The dataset has a ratio of 0.65/0.10/0.25 for train/val/test. Target network is TextCNN \cite{TextCNN}. The hyperparameters for the \ourmodel model are $p=16, \lambda=4$. Test rays is 25 evenly preference vectors. The reference point for hypervolume is (2, 2). 

In this case, \ourmodel has a greater hypervolume than previous techniques and still permits the Pareto front to move deeper. For COSMOS, due to the mapping from a preference vector $r$, to the huge dimensionality of the embedding feature, it does not converge.

\subsection{Multi-Output Regression}
To demonstrate the viability of our strategy in high-dimensional space, we conduct experiments on 2 datasets:

\indent - \textbf{Jura} \cite{Goovaerts1997}: In this experiment, the goal variables are zinc, cadmium, copper, and lead (4 tasks), whereas the predictive features are the other metals, the type of land use, the type of rock, and the position coordinates at 359 different locations. The dataset has a ratio of 0.65/0.15/0.20 for train/val/test.

\indent - \textbf{SARCOS} \cite{Sethu2000}:  The goal is predict pertinent 7 joint torques (7 tasks) from a 21-dimensional input space (7 joint locations, 7 joint velocities, 7 joint accelerations). There are 4449 and 44484 examples on testing/training set. As validation set, 10\% of the training data are used. 

The target network in both experiments is a Multi-Layer Perceptron with 4 hidden layers containing 256 units. We set $p=8, \lambda=0.001$ for \ourmodel on both two datasets. The reference point for calculating HV is $(1, 1, \dots, 1)$. \ourmodel outperforms all other baselines in terms of Hypervolume. 

\subsection{Ablation Study}
\begin{figure}[!htb]
  \centering
  \includegraphics[width=0.75\columnwidth]{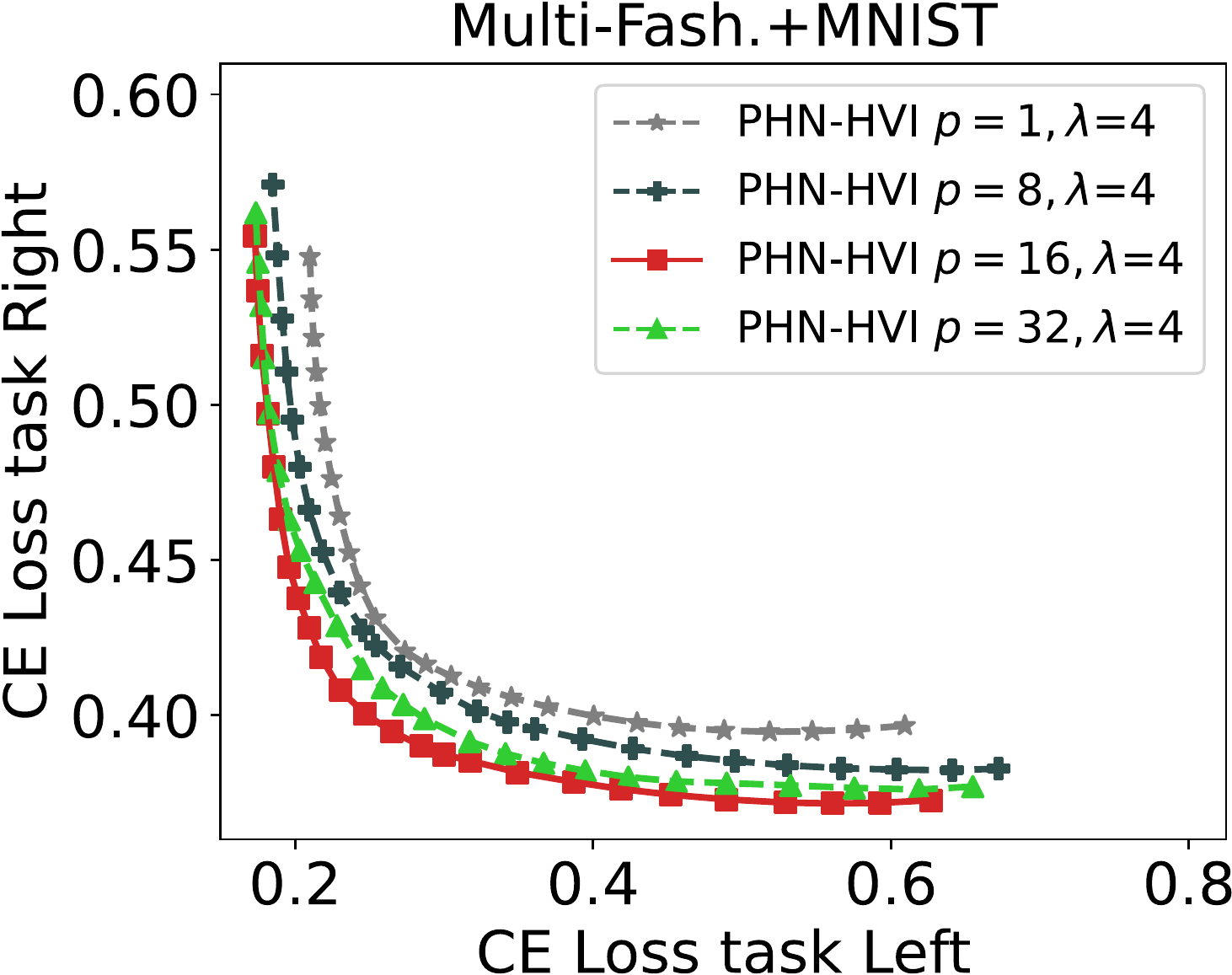}
  \caption{Performance of \ourmodel when $p$ varies}
  \label{fig:head}
\end{figure}

\begin{figure}[!htb]
  \centering
  \includegraphics[width=0.75\columnwidth]{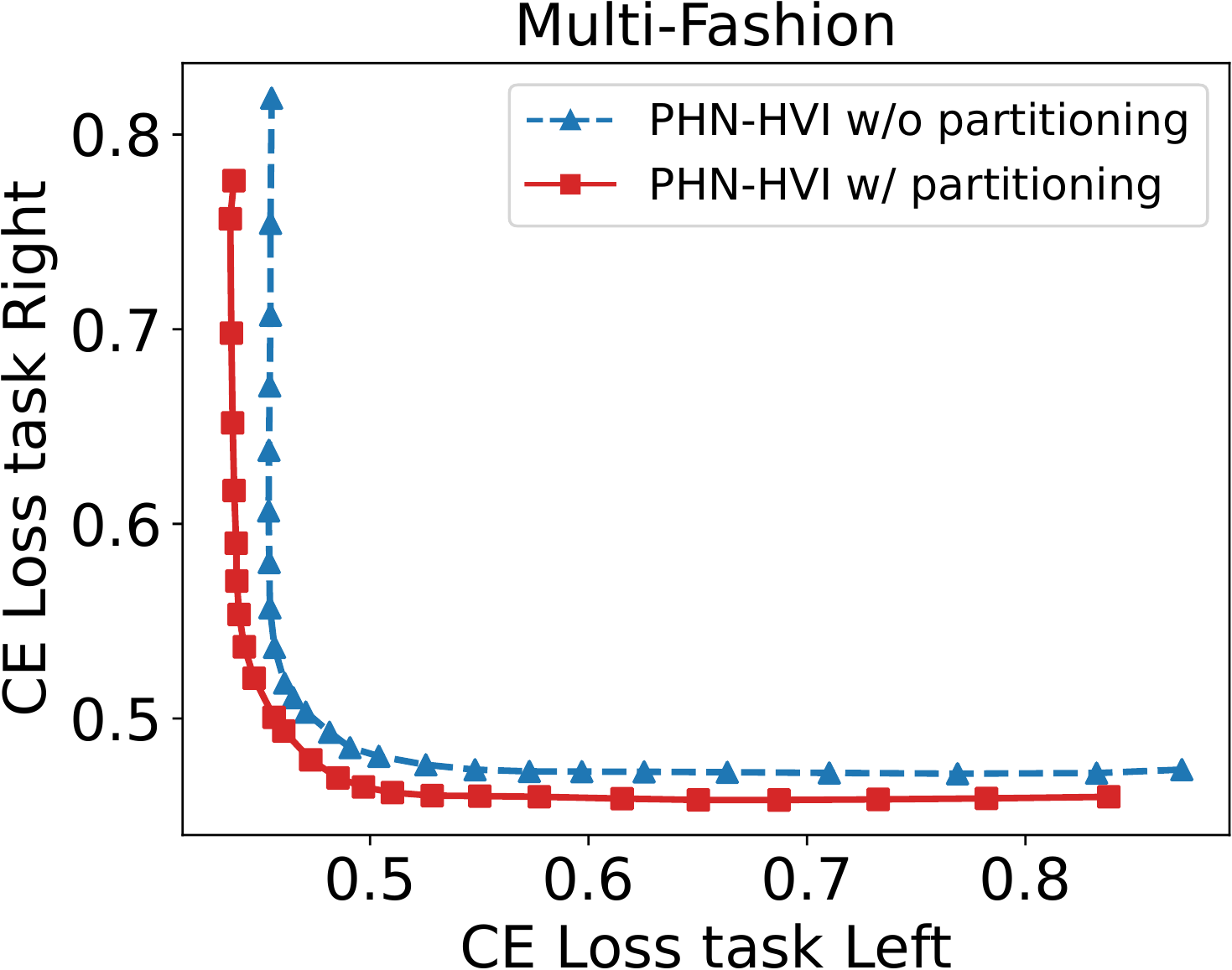}
  \caption{The effect of loss space partitioning}
  \label{fig:adap_correct}
\end{figure}

\textbf{Number of Rays $p$.} Figure \ref{fig:head} demonstrates that the quality of the Pareto front increases with the number of rays, but up to a certain point, adding more rays no longer significantly improves the results. That means our framework doesn't require too many sampling rays to get a good performance. 
\indent \textbf{Partition.} As shown in Figure \ref{fig:adap_correct}, partitioning algorithm makes it easier for the cosine similarity function and the HV function to cooperate and enhances \ourmodel performance.

\begin{figure}[!htb]
\centering
\includegraphics[width=0.75\columnwidth]{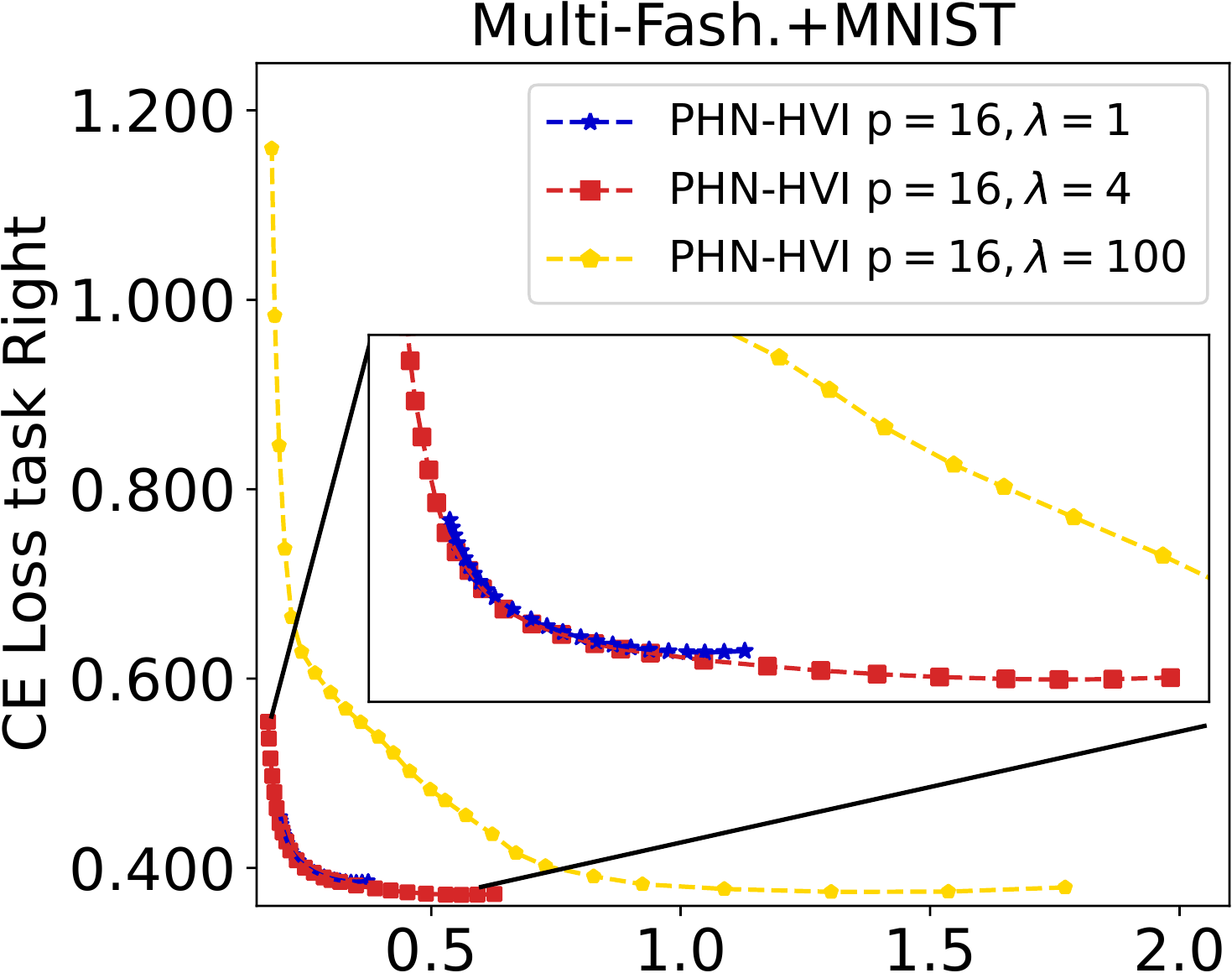} \\
\vspace{5pt}
\includegraphics[width=0.75\columnwidth]{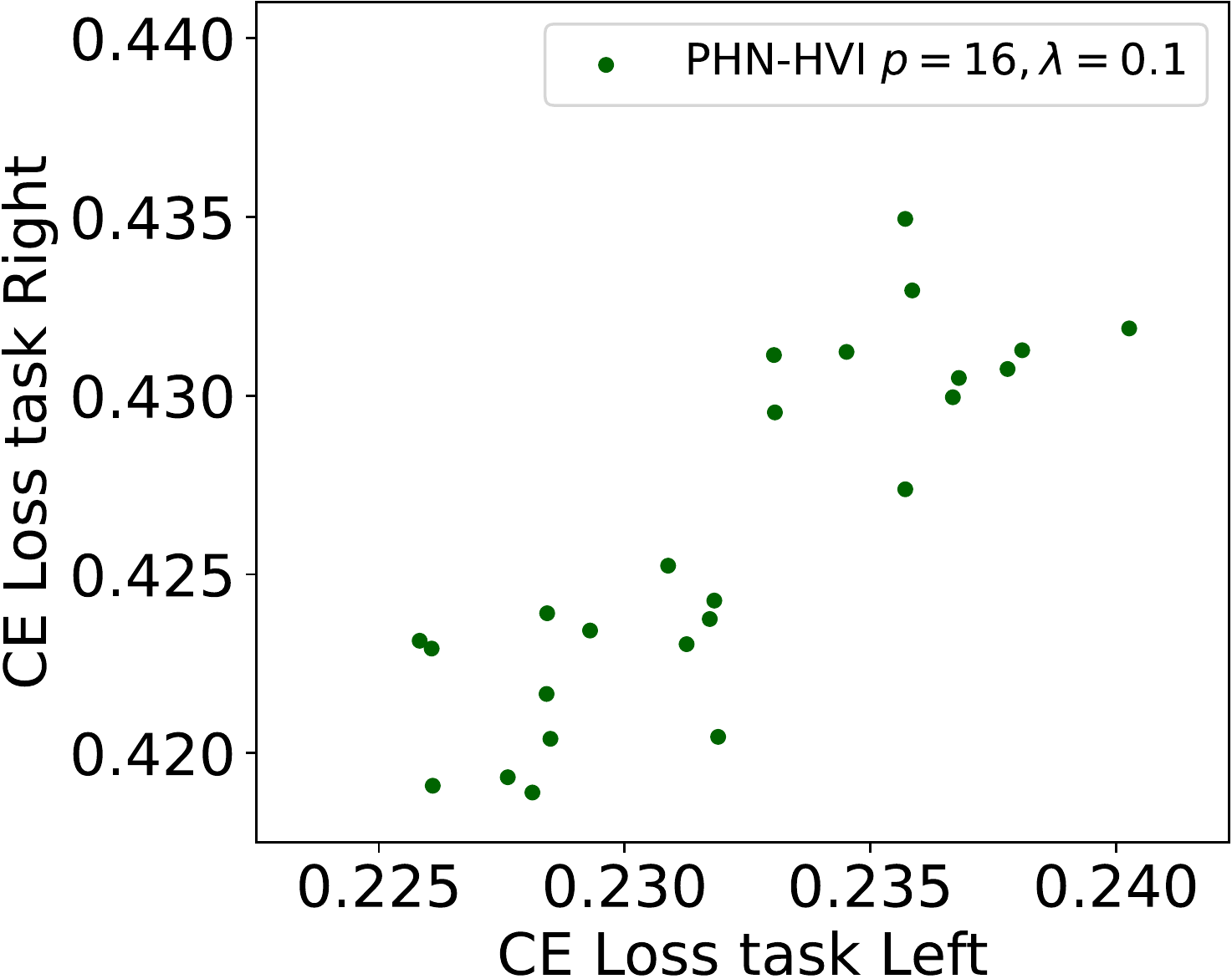}
\caption{The impact of hyperpareter $\lambda$}
\label{eq:lambda}
\end{figure}

\textbf{Cosine Similarity.} The cosine similarity function is critical in the convergence of \ourmodel and helps in the spread of the Pareto Front. In Figure \ref{eq:lambda}, if $\lambda$ is very large ($\lambda=100$), Pareto Front is very widely dispersed, but it is quite shallow. If $\lambda$ is very small $(\lambda = 0.1)$, \ourmodel can't generate Pareto Front. Therefore, selecting a suitable lambda that balances the HV function and the cosine similarity function is critical for the \ourmodel to work effectively. 

\section{Conclusion and Future Work}
\label{st:6}
In this paper, we propose \ourmodel with Multi-Sample Hypernetwork, which utilizes a variety of trade-off vectors simultaneously, followed by hypervolume maximization to improve the PFL problem. This approach also opens up a wide range of potential research directions. On one hand, it is necessary to investigate theoretically for which objective functions the hypernetwork-based PFL methods will guarantee the convergence. On the other hand, it is shown that hypernetwork-based PFL can not approximate well disconnected-Pareto fronts. Hence, the question of whether PFL may be solved effectively without hypernetwork is very crucial to consider.  

\appendix

\section{Experimental Details}
The following parameters of PHN-LS, PHN-EPO, and COSMOS have been best selected based on references from the original papers \citep{navon2021learning, ruchte2021scalable}. For the hypernetwork, we use an MLP with 2 hidden layers and linear heads. We set the hidden dimension to 100. We use dropout at the last layer of the hypernetwork to avoid overfitting. Adam is the optimizer for Multi-Task Learning tasks.

\textbf{Image Classification:} We train all methods using an Adam optimizer with learning rate $5e^{-4}$ with \ourmodel,  $1e^{-3}$ with otherwise for maximal 400 epochs and batch size 256. If the algorithm fails to update the best model for 10 consecutive epochs, we multiply the learning rate by $\sqrt{2}$ and terminate the program early if 35 consecutive epochs. With PHN-LS and PHN-EPO, $\alpha = 0.2$. With COSMOS, $\alpha=1.2$ and $\lambda = 2/2/8$ correspoding to Multi-MNIST, Multi-Fashion, Multi-Fashion+MNIST. $\alpha = (0.5, 0.5)$ for \ourmodel for three datasets. The dropout ratio for \ourmodel is 0.1 and 0 for other methods.

\textbf{Text Classification and Regression:} We train all methods using an Adam optimizer with learning rate  $1e^{-3}$ for maximal 400 epochs and batch size 256. If the algorithm fails to update the best model for 10 consecutive epochs, we multiply the learning rate by $\sqrt{2}$ and terminate the program early if 35 consecutive epochs. $\alpha = 0.2$ for PHN-LS and PHN-EPO, $\alpha = (0.5, 0.5)$ for \ourmodel. COSMOS doesn't converge on this task. The dropout ratio is 0.1 for all methods.

\textbf{Multi-Output Regression:} We train all methods using an Adam optimizer with learning rate  $1e^{-3}$ for maximal 1000 epochs and batch size 64/512 for Jura/SARCOS. If the algorithm fails to update the best model for 40 consecutive epochs, we multiply the learning rate by $\sqrt{2}$. We set $\alpha = 0.2$ for PHN-LS, PHN-EPO and COSMOS, $\alpha = (1/7,\dots, 1/7)$ for \ourmodel.  $\lambda=1$ for COSMOS. We normalize output variables by dividing them by the maximal value on the Jura dataset and the quantile of 0.9 on the SARCOS dataset. The dropout ratio is 0. for all methods. Word embedding matrix is pre-train matrix\footnote{Available at \url{https://nlp.stanford.edu/projects/glove/}}  glove.840B.300d which is unsupervised trained by Stanford research group.

\section{Warm up for Multi-Sample Hypernetwork}
We warm up Multi-Sample Hypernetwork by optimize the follow problem by 1 epoch to help quality function $ Q(\mathfrak{L}(\Theta,x, y))$ and extra criterion function interact well:
\begin{align}
    \min_\phi - \mathbb{E}_{r^i\sim \text{Dir}\left(\alpha \right), (x,y) \sim p_D} \sum_{i=i}^p F(\mathcal{L}^i, r^i)
\end{align}

\section{Additional results of Partition}
\begin{figure}[htb]
    \centering
    \includegraphics[width=.95\columnwidth]{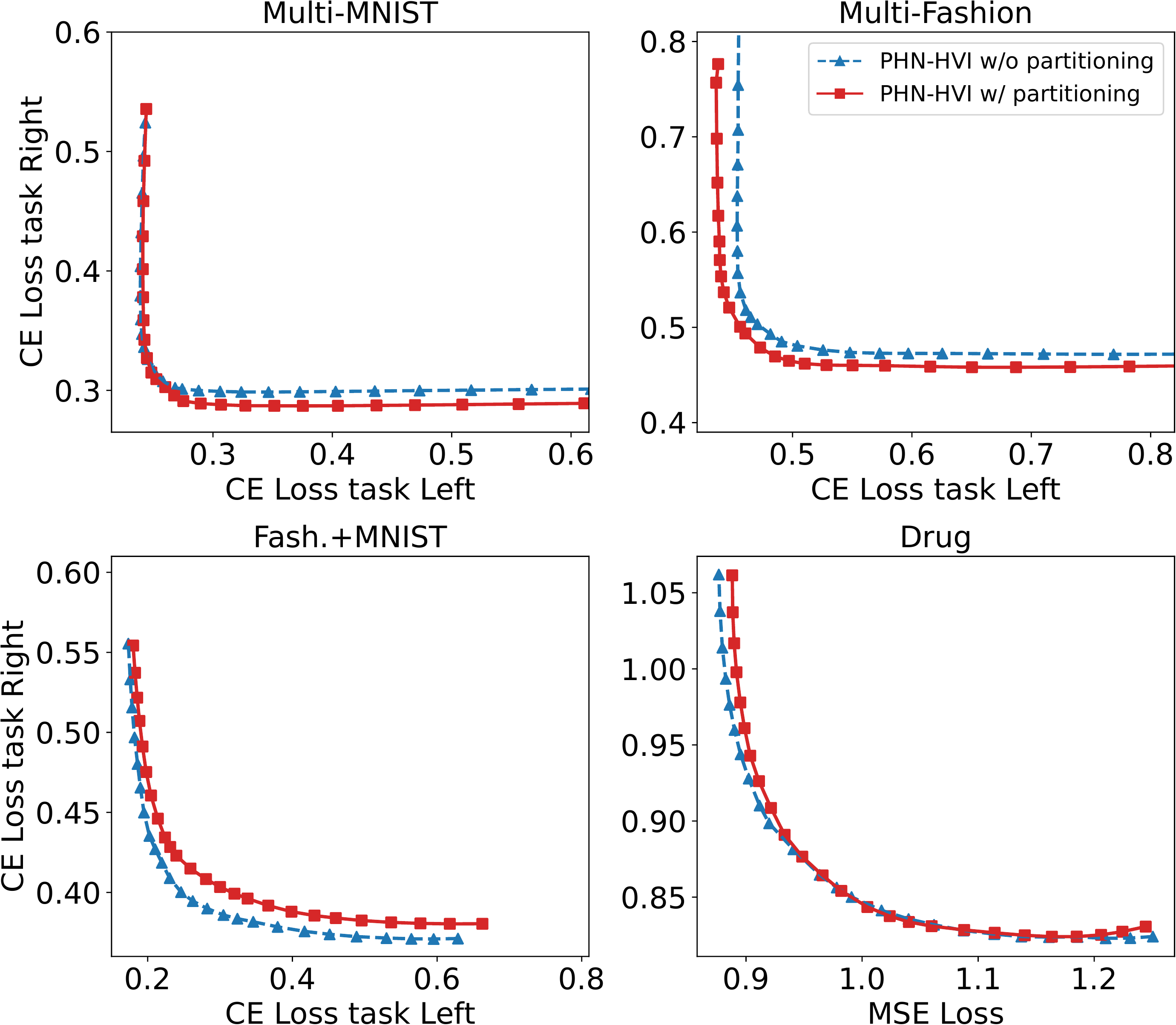}
    \caption{Impact of partition algorithm in different datasets}
    \label{fig:partition_all}
\end{figure}
As Figure \ref{fig:partition_all}, the partition sampling algorithm works efficiently in some particular case such as Multi-Fashion. Because it doesn't always have a Pareto optimal solution along to a specific preference ray, the way partition algorithm enforces sampling rays cover the entire space maybe have some negative effect, especially in Multi-Fashion+MNIST, in which the loss functions have a different about magnitude.

\section{Incompatibility of LS and Cosine Similarity }
\begin{figure}[!htb]
    \centering
    \includegraphics[width = .65\columnwidth]{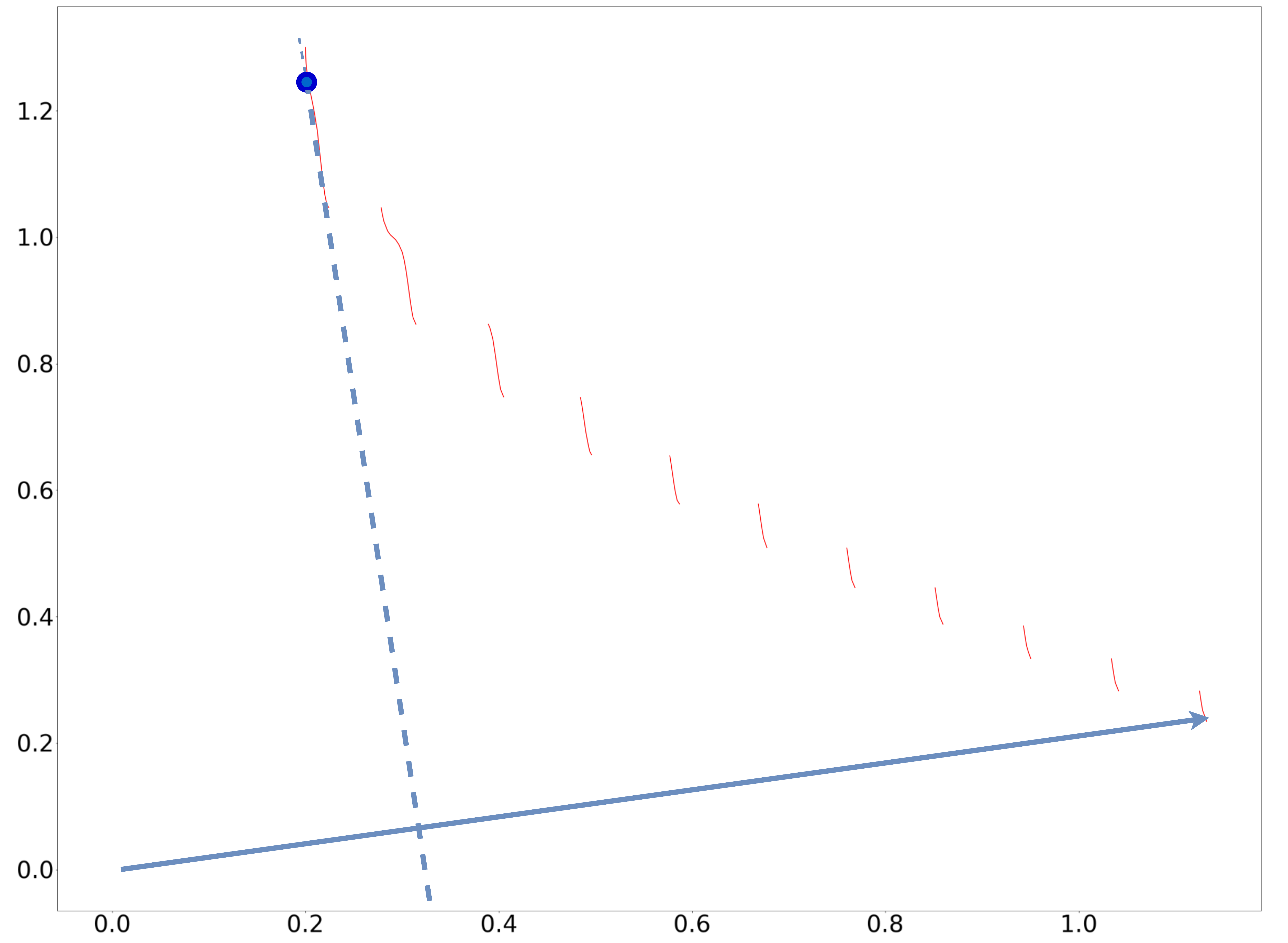}
    \caption{LS vs Cosine similarity}
    \label{fig:ls_cs}
\end{figure}

In Problem 3 of Toy Examples, COSMOS is not converged because the solution of Linear Scarlarization (LS) is very different from a ray. In Figure \ref{fig:ls_cs}, the blue point is the corresponding solution of LS with the given preference ray.

On the other hand, because \ourmodel generates multiple points spread evenly in every iteration, each point is close to a preference ray, so the HV function and cosine similarity have an effective interaction.

\section{Training cost}
At each iteration, our approach requires running and storing $p$ target networks. Even though the training cost is more expensive, the forward computation of the p networks can be performed in parallel. As shown in Figure \ref{fig:head}, up to a certain point, adding more rays no longer significantly improves, we don't require a large number of target networks $p$ to produce good results, therefore the cost is acceptable. It is important to note that our hypernetwork's scale is the same as that of other baselines like PHN-LS and PHN-EPO. Without extra trainable parameters, our proposed method still has a non-trivial advancement when compared to other methods, which can’t improve the result with an increase in running time. At the testing phase, the inference time of Multi-Sample Hypernetwork and other baselines are identical, because it simply requires a preference ray to provide a corresponding target network which is a Pareto optimal solution.

\section{Selecting the HV reference point}
The J-dimensional reference point r is a hyperparameter in hypervolume maximization. When we don't know the true Pareto front and don't prioritize any objectives, it should be fixed in every iteration to control the change in hypervolume and should be equal on all coordinates. Fortunately, selecting a reference point such that well-spread predictions can approximate the entire Pareto front is not hard. Choosing a reference point with higher coordinates than random initialization of network losses is frequently sufficient. In some special early iterations, if the current losses are larger than the reference point, we will rescale the current losses by a factor $\gamma=1.1$ as a temporary reference point.

\section*{Acknowledgments}
This research is supported by VinUniversity under its Fast-Track funding Programme (Proposal No. 22-0003-P0001). The authors also gratefully acknowledge VinUni-Illinois Smart Health Center, VinUniversity for providing GPUs resource for this research.     



\bibliography{aaai23.bib}

\end{document}